\documentclass[runningheads]{llncs}
\usepackage[T1]{fontenc}
\usepackage{amsmath}
\usepackage{svg}
\usepackage{multirow}
\usepackage{graphicx}

\begin{document}
\title{Education Distillation: Let the Model Learn in the School}
\makeatletter
\author{Ling Feng\inst{1} \and
Tianhao Wu\inst{1} \and
Xiangrong Ren\inst{1} \and
Zhi Jing\inst{1} \and
Xuliang Duan\inst{1}
}

\authorrunning{F. Author et al.}
\institute{Sichuan Agricultural University, Sichuan, China\\
\email{202105857@stu.sicau.edu.cn, 202205793@stu.sicau.edu.cn, 202204704@stu.sicau.edu.cn, 202308498@stu.sicau.edu.cn, duanxuliang@sicau.edu.cn}
}
\makeatother
\maketitle              % typeset the header of the contributiond
%e
\begin{abstract}
This paper introduces a new knowledge distillation method, called education distillation (ED), which is inspired by the structured and progressive nature of human learning. ED mimics the educational stages of primary school, middle school, and university and designs teaching reference blocks. The student model is split into a main body and multiple teaching reference blocks to learn from teachers step by step. This promotes efficient knowledge distillation while maintaining the architecture of the student model. Experimental results on the CIFAR100, Tiny Imagenet, Caltech and Food-101 datasets show that the teaching reference blocks can effectively avoid the problem of forgetting. Compared with conventional single-teacher and multi-teacher knowledge distillation methods, ED significantly improves the accuracy and generalization ability of the student model. These findings highlight the potential of ED to improve model performance across different architectures and datasets, indicating its value in various deep learning scenarios. Code examples can be obtained at: https://github.com/Revolutioner1/ED.git.

\end{abstract}
\section{Introduction}n
Human education progresses gradually from primary school to middle school and then to university. What about Convolutional Neural Networks? Knowledge distillation (KD), first proposed by Hinton, is a model compression technique based on the "teacher-student network concept". The goal of traditional knowledge distillation is to transfer knowledge from a complex network (teacher network) to a simple network (student network) by minimizing the KL divergence between the softened outputs of the two networks\cite{ref_article1}.

The field of knowledge distillation is constantly evolving. Many studies explore new ways of knowledge transfer, such as starting from output layers, intermediate layers, attention maps, and other aspects. Zagoruyko S et al. proposed a method to transfer attention maps from the teacher Convolutional Neural Network to the student network to improve the performance of the student network. This represents a new breakthrough and development compared to traditional knowledge distillation methods\cite{ref_lncs1}. Ahn S proposed to transfer knowledge by maximizing the mutual information between the teacher and student networks, providing new ideas and methods for improving model performance\cite{ref_lncs2}. Romero A et al. proposed using the representations of the intermediate layers of the teacher network as hints to help train a student network that is deeper and narrower than the teacher network, opening up a new path for solving the problem of model compression and enhancing the performance of the student network\cite{ref_lncs3}. Tian Y proposed using a family of contrastive objectives to capture correlations and high - order output correlations. This innovatively applies the idea of contrastive learning to knowledge distillation, enabling the extraction of knowledge from one neural network to another\cite{ref_lncs4}.

Similarly, predecessors have proven that multi - teacher knowledge distillation methods also perform excellently. Multi - teacher knowledge distillation improves the effectiveness of distillation by integrating the predictions of multiple teachers, and there are already several representative methods. For example, Shan You et al. combined the knowledge of the output layers and intermediate layers of different teacher networks and proposed a voting strategy to unify the relative difference information from multiple teachers\cite{ref_lncs5}. Kisoo Kwon et al. introduced an entropy - based knowledge distillation method\cite{ref_lncs6}. Shangchen Du et al. proposed an adaptive ensemble knowledge distillation method, which uses the gradient information of multiple teacher models and introduces a tolerance parameter to dynamically adjust the weights of teacher models\cite{ref_lncs7}. Hailin Zhang et al. proposed a knowledge distillation method that dynamically adjusts the weights of multi - teacher models by combining the confidence of teacher model predictions with true labels\cite{ref_lncs8}.

Undoubtedly, the above - mentioned methods are all excellent. However, we prefer to approach distillation in a more intuitive way, enabling the model to progress step by step. Education requires many teachers. We can envision: If we combine multi - teacher knowledge distillation with education, can Convolutional Neural Networks also develop from the "primary school" stage to the "university" stage just like humans? To this end, we propose an education distillation method. Through incremental learning and model pruning, the student model can gradually improve from the "primary school" level to the "university" level, thus absorbing knowledge more effectively.

Meanwhile, we introduce the concept of "reference books" and design a teaching reference block to further optimize the learning process without changing the final architecture of the student model. Compared with the five single - teacher knowledge distillation methods and four multi - teacher knowledge distillation methods mentioned above, education distillation significantly improves the overall performance of the teacher - student architecture.
\begin{figure}[htbp]
\centering
\includegraphics[width=0.6\textwidth]{11.pdf}
\caption{In educational distillation, the student model is divided into a main body and three teaching reference blocks. At the same time, the dataset is partitioned into three sub-datasets for distillation. The model starts at the "primary school" stage, where it trains on sub-dataset 1. When the $n$-th epoch is reached and fitting is achieved, the second teaching reference block is added, and the model progresses to the "secondary school" stage, training on sub-dataset 2. When the $m$-th epoch is reached and fitting is achieved, the final reference block is added, and the model reaches the "university" stage, training on sub-dataset 3.  }
\label{fig1}
\end{figure}
\section{Method}
\subsection{Education Distillation Method}
Before delving into the concept of education distillation, let's first consider a learning scenario. Suppose there is a student who needs to gradually master three subjects - arithmetic, equations, and calculus - in primary school, middle school, and university. In primary school, the student starts with arithmetic. As learning progresses, new teachers will gradually teach equations. In university, the student begins to learn calculus, integrating all the accumulated mathematical knowledge. Can training a student model in a similar step - by - step learning manner improve the effectiveness of model training?

Convolutional Neural Networks, such as ResNet\cite{ref_lncs9}, WRN\cite{ref_lncs10}, ShuffleNet\cite{ref_lncs11}, VGG\cite{ref_lncs12}, and MobileNetV2\cite{ref_lncs13}, are generally composed of multiple building blocks, which are called "blocks" or "layers". These blocks are stacked together to form the entire network framework. This modular design not only enables Convolutional Neural Networks to effectively learn hierarchical features from data, but also allows them to adapt to tasks of different complexities.

In the process of education distillation, we split the student model into a main body and multiple reference layers. Then, we divide the dataset into subsets according to the number of reference layers. During training, every time a new module is added, a subset is introduced. This process continues until the complete student model is fully restored.

Let us take the three educational stages of primary school, middle school, and university as an example. The distillation process of education for refany convolutional network can be divided into the learning stages shown in Fig.\ref{fig1}. 

In the process depicted in Fig.\ref{fig1}, the main body retains the encoder of the student network, and the designed teaching reference block (TR Block) serves as the decoder. The complete student network is divided into one main body and three teaching reference blocks for distillation.

At the start of training (corresponding to the primary school stage), the main body adds a teaching reference block to become a new main body and learns from Sub - dataset 1. In the second stage (representing the middle school stage), the main body adds another teaching reference block to form a new main body, and simultaneously learns from Sub - dataset 2 and Sub - dataset 1. Finally, in the third stage (similar to the university stage), the main body adds the last teaching reference block to restore the complete student model and learns the content of the entire dataset.

At each stage, as the modules are allocated, the teaching reference blocks are also updated synchronously to assist in training the student network. By gradually integrating these modules, the student model is ultimately reconstructed into a complete network. Instead of learning the entire data at once, it learns the content of the sub - datasets one by one. Each sub - dataset is trained by a dedicated teacher model for the corresponding stage.

\subsection{The Phenomenon of Catastrophic Forgetting}
During the process of a model learning a new task, a remarkable phenomenon occurs. That is, the model's grasp of knowledge and performance on previously learned old tasks decline significantly. This means that while the model is striving to adapt to new data, it seems to "forget" the key information learned before. This is the so - called phenomenon of catastrophic forgetting\cite{ref_lncs16}.

In the process of education distillation, we introduce the teaching reference block to avoid the problem where the model only focuses on learning a sub - dataset and forgets the content of other sub - datasets. The teaching reference block consists of one block of the student network, a 1×1 convolutional layer, and a specially designed adaptive average pooling layer.

Whenever the training of a sub - dataset reaches convergence and a new teaching reference block is added to the main body, we remove the previous convolutional layer and pooling layer and freeze the previous parameters. This ensures that each actual component of the student network corresponds to a specific sub - dataset. The teaching reference block mainly acts as an auxiliary component of the student network during the education distillation process. Therefore, during the last update of the main body, the teaching reference block will be restored to the original decoder part of the student network.

This approach not only strengthens the training process but also effectively guarantees the integrity of the network structure. Final experiments show that the design of such a teaching reference block can indeed significantly improve the accuracy of the model.

 \subsection{Formalization}
 The whole process is analyzed as follows: $M_t(\cdot)$ indicates the learned model of the student model at incremental stage $t$, the classifier $g(x)$, the original student model $S$, and the teaching reference blocks \{$f_l(\cdot)$; $l =1,2,3,..., n$\}, which are expressed as follows:
\begin{equation}
M_t(x)=g\left(h_t\right)=g\left(S \circ f_1 \circ f_2 \circ f_3 \circ \ldots \circ f_t(x)\right)
\end{equation}

Notably $h_t$ is the input model eigenvector, represented each sub-dataset and originating from the dataset $H$  , $h_t \in H$, which is denoted by
\begin{equation}
h_1 \cup h_2 \cup h_{3 \ldots} \cup h_t=H
\end{equation}
$h_t$ with the corresponding $f_t(\cdot)$ are given the corresponding mapping result
\begin{equation}
Z_{t, h}=\left\{Z_{f_1, 1}, Z_{f_1, 2}, \ldots, Z_{f_{t, h}}\right\}
\end{equation}

For the teacher model, $T_t(\cdot)$ denotes the set of all $t$ teacher models, and each group $h_t$ will have a uniquely mapped teacher model $T_t(\cdot)$, which is expressed as:
\begin{equation}
T_t(x)=g\left(h_t\right)=g\left(T_1(x) \cup T_2(x) \ldots \cup T_t(x)\right)
\end{equation}

Eventually both ht and the corresponding $T_t$ will get the corresponding mapping result
\begin{equation}
G_{t, h}=\left\{G_{T_1, 1}, G_{T_2, 2}, \ldots, G_{T_t, h}\right\}
\end{equation}
$\mathcal{L}_h(\cdot)$ is the loss for a particular sub-dataset. In training $M_t(\cdot)$, the distillation loss under the sub-dataset is
\begin{equation}
\mathcal{L}_h(Z_{t,h},G_{t,h}) = 
\sum_{Z_i\in Z_{t,h}} \sum_{G_i\in G_{t,h}} 
KL\left(\mathrm{softmax}\left(\frac{Z_i}{\tau}\right),\mathrm{softmax}\left(\frac{G_i}{\tau}\right)\right)
\end{equation}
where $KL(\cdot)$ denotes the KL dispersion and $\tau$ denotes the distillation temperature.

For $M_t(\cdot)$ training, the sum of all task losses can then be expressed as:
\begin{equation}
\mathcal{L}_h\left(Z_t, G_t, y\right)=\alpha * \mathcal{L}_{K D}\left(Z_t, G_t\right)+(1-\alpha) * \mathcal{L}\left(Z_t, y\right)
\end{equation}
where $y$ denotes the true label of the input eigenvector and $\alpha$ denotes the weight of the distillation loss.

Then the loss approximation for the student network is calculated as follows
\begin{equation}
\begin{aligned}
ED_{LOSS} = \mathcal{L}_H(Z_t,G_t,y) 
\approx \mathcal{L}_{h_1}(Z_{t,h_1},G_{t,h_1},y) + 
\mathcal{L}_{h_2}(Z_{t,h_2},G_{t,h_2},y) + \\
\mathcal{L}_{h_3}(Z_{t,h_3},G_{t,h_3},y) + \cdots
\end{aligned}
\end{equation}
$U_t$ is the feature space corresponding to each set of $h_t$, which is denoted as:
\begin{equation}
U_1 \cup U_2 \cup U_3 \ldots \cup U_n \ldots \cup U_t=U
\end{equation}
\begin{equation}
U_1 \cap U_2 \cap U_3 \ldots \cap U_n \ldots \cap U_t=\emptyset
\end{equation}

It is inefficient for a small model to learn directly from the full feature space $U$. However, in education distillation, $M_t(\cdot)$ begins by learning from a smaller feature space $U_1$. As the number of incremental basic blocks(teaching reference blocks) $f_l(\cdot)$ increase, the small feature space gradually expands into a larger feature space. Additionally, there is no overlap between the small feature space $U_n$ and the newly expanded feature space $U_t$, thus improving the model's efficiency in learning features.

\section{Experiments}
In this section, to prove the effectiveness of the proposed education distillation (ED) method, we conducted experiments on the CIFAR100 dataset\cite{ref_lncs14} using nine different teacher - student architectures. We compared our method with single - teacher knowledge distillation methods and multi - teacher knowledge distillation methods. In the multi - teacher knowledge distillation methods, three teacher models were used in all cases. These three teacher models exhibited an average top - 1 accuracy value.

\begin{table}
\label{1}
\centering
\caption{Top-1 test accuracy of ED methods by distilling the knowledge on single-teachers with the same architectures.}\label{tab1}
\begin{tabular}{|c|c|c|c|c|c|}
\hline
Model&  WRN40\_2 & VGG13 & \multicolumn{2}{c|}{ResNet110}& ResNet56\\
Teacher & 76.76 & 75.22 & \multicolumn{2}{c|}{74.26} & 73.47 \\
Student &  WRN16\_2 & VGG8 & ResNet32 & ResNet20 & ResNet20\\
\hline
KD\cite{ref_article1}& 75.28 & 73.62 & 74.09 & 71.27 & 71.08  \\
AT\cite{ref_lncs1} & 75.34  &  73.27 & 73.31 & 71.26 & 70.13\\
VID\cite{ref_lncs2} & 75.54 &  73.28 & 73.54 & 70.99 & 70.7\\
FitNet\cite{ref_lncs3} & 75.44 & 73.4 & 73.37 & 71.03 & 71.19  \\
CRD\cite{ref_lncs4} & 75.38 & 73.41  & 73.54 & 70.72 & 70.98\\
\textbf{ED} & \textbf{77.25} & \textbf{75.5} & \textbf{74.47} & \textbf{72.33} & \textbf{73.85}\\
\hline
\end{tabular}
\end{table}
\begin{table}
\centering
\caption{Top-1 test accuracy of ED methods by distilling the knowledge on multiple-teachers with the same architectures.}\label{tab2}
\begin{tabular}{|c|c|c|c|c|c|}
\hline
Model&  WRN40\_2 & VGG13 & \multicolumn{2}{c|}{ResNet110}& ResNet56\\
Teacher & 76.76 & 75.22 & \multicolumn{2}{c|}{74.26} & 73.47\\
Student &  WRN16\_2 & VGG8 & ResNet32 & ResNet20 & ResNet20\\
\hline
AVER\cite{ref_lncs5} & 76.17 & 74.21  & 75.19 &  71.38& 71.2  \\
EBKD\cite{ref_lncs6} & 76.33  & 74.08  & 73.97 & 71.5 & 71.18\\
AEKD\cite{ref_lncs7} & 75.99 &  74.1 & 74.31 & 71.67 & 71.23\\
CAMKD\cite{ref_lncs8} & 76.27 & 74.4 & 74.32 & 71.27 & 70.79  \\
\textbf{ED} & \textbf{77.25} & \textbf{75.5} & \textbf{74.47} & \textbf{72.33} & \textbf{73.85}\\
\hline
\end{tabular}
\end{table}
\begin{table}
\centering
\caption{Top-1 test accuracy of ED methods by distilling the knowledge on single-teachers with the heterogeneous architectures.}\label{tab3}
\begin{tabular}{|c|c|c|c|c|c|}
\hline
Model&  WRN40\_2 & VGG13 & ResNet56 & ResNet56 \\
Teacher & 76.76  & 75.22 & 74.26 & 73.47  \\
Student &  ShuffleNetV1 & MobileNetV2 & MobileNetV2 & VGG8 \\
\hline
KD\cite{ref_article1} & 75.82 & 68.02  & 68.43 & 73.52 \\
AT\cite{ref_lncs1} &  75.43 & 68.39  & 69.65 & 73.23 \\
VID\cite{ref_lncs2} & 75.76 &  68.43 & 68.04 & 73.46 \\
FitNet\cite{ref_lncs3} & 75.75 & 67.93 & 69.32 &  73.57  \\
CRD\cite{ref_lncs4}& 75.96 & 68.27  & 69.44 & 73.27 \\
\textbf{ED} & \textbf{77.63} & \textbf{70.05} & \textbf{72.33} & \textbf{76.78}  \\
\hline
\end{tabular}
\end{table}
\begin{table}
\centering
\caption{Top-1 test accuracy of ED methods by distilling the knowledge on multiple -teachers with the heterogeneous architectures.}\label{tab4}
\begin{tabular}{|c|c|c|c|c|c|}
\hline
Teacher&  WRN40\_2 & VGG13 & ResNet56 & ResNet56 \\
Teacher & 76.76  & 75.22 & 74.26 & 73.47  \\
Student &  ShuffleNetV1 & MobileNetV2 & MobileNetV2 & VGG8 \\
\hline
AVER\cite{ref_lncs5} & 78.08 & 69.93 & 71.38 &  75.81  \\
EBKD\cite{ref_lncs6} & 77.84  & 69.55 & 71.5 & 75.51  \\
AEKD\cite{ref_lncs7} & 77.56 &  69.35 & 71.67 & 75.61  \\
CAMKD\cite{ref_lncs8} & 77.02 & 69.22 & 71.27 & 75.57    \\
\textbf{ED}& \textbf{77.63} & \textbf{70.05} & \textbf{72.33} & \textbf{76.78}  \\
\hline
\end{tabular}
\end{table}

\textbf{Hyperparameters}: All neural networks were optimized using Stochastic Gradient Descent (SGD) with a momentum of 0.9 and a weight decay of 0.0001. The batch size was set to 32, and the initial learning rate was set to 0.05. During a total of 240 training epochs, the learning rate was multiplied by 0.1 at epochs 150, 180, and 210. The temperature T for all methods was set to 4, and $\alpha$ was set to 0.3. In epochs 50, 80, and 210, we added building blocks to the student network and updated the teaching reference layers. Meanwhile, the sub - datasets were divided in a 1:1:1 ratio for training.

\textbf{Comparison of Isomorphic Teacher-Student Models}: In the network of isomorphic teacher-student models, Table \ref{tab1} and Table \ref{tab2} show the comparison of top-1 accuracies of single-teacher and multi-teacher knowledge distillation methods on the CIFAR100 dataset. We found that the education distillation (ED) method outperforms all its competitors across various architectures.

For example, in single-teacher knowledge distillation, when using WRN40\_2 as the teacher model and WRN16\_2 as the student model, the ED method achieves an average accuracy improvement of 1.854\%. In multi-teacher knowledge distillation, when using ResNet56 as the teacher model and ResNet20 as the student model, the ED method increases the average accuracy by 2.2\%.

Overall, the ED method generally performs better in teacher-student configurations with the same architecture. Carefully selecting the distillation strategy and the combination of teacher-student architectures can maximize the learning effect of the student model.

\textbf{Comparison of Heterogeneous Teacher-Student Models}: In the network of heterogeneous teacher-student models, as shown in Table \ref{tab3} and Table \ref{tab4}, the ED method also demonstrates strong performance across various architectures.

Specifically, whether using WRN40, VGG13, ResNet56, or another ResNet56 as the teacher network, the ED method effectively enhances the performance of student networks such as ShuffleV1, MobileNetV2, and VGG8.

For example,when WRN40 is used as the teacher network, uusinge ED method, the ShuffleV1 student network can achieve an accuracy of 77.63\%, and the accuracy of MobileNetV2 can reach up to 72.33\% at its best. This indicates that the ED method can effectively transfer knowledge between different network architectures.

\begin{table}
\centering
\caption{For isomorphic models, it refers to the change in the accuracy of each sub-data when adding a teaching reference block each time. Experiments show that after adding a teaching reference block, the model does not forget the content of the previous sub-dataset. At the same time, as the model approaches completion, the accuracy gradually improves. }\label{tab5}
\begin{tabular}{|c|c|c|c|}
\hline
ResNet56/ResNet20& Sub-dataset 1 & Sub-dataset 2 &Sub-dataset 3\\
\hline
Primary & 45.64 &\slash &\slash\\
Middle& 61.44& 60.56&\slash\\
University & 75.33&73.96& 71.86\\
\hline
\end{tabular}
\end{table}

\begin{table}
\centering
\caption{For heterogeneous models, they exhibit the same performance as isomorphic models. After adding a teaching reference block, the model does not forget the content of the previous sub-dataset. Meanwhile, as the model becomes more complete, the accuracy gradually increases.  }\label{tab6}
\begin{tabular}{|c|c|c|c|}
\hline
ResNet56/MobileNet2& Sub-dataset 1 & Sub-dataset 2 &Sub-dataset 3\\
\hline
Primary & 43.64 &\slash & \slash\\
Middle& 63.67& 60.56&\slash\\
University & 73.22&72.78& 70.99\\
\hline
\end{tabular}
\end{table}

\textbf{The Phenomenon of Catastrophic Forgetting}:As shown in Table \ref{tab5} Table \ref{tab6}, each time a new teaching reference block and a sub-dataset are incrementally added to the main body, the learning of previous sub-datasets is not forgotten. Meanwhile, due to the improvement of the model's performance, the training accuracy is also increasing. The teaching reference block can indeed solve the problem of catastrophic forgetting.
n
\textbf{Performance on Tiny Imagenet}: Finally, as shown in Table \ref{tab7} Table \ref{tab8}, to further verify the generalization ability of the education distillation method, we applied it to the Tiny Imagenet\cite{ref_lncs15}, Caltech\cite{ref_lncs17} and Food-101\cite{ref_lncs18} datasets and observed excellent performance results.

It should be noted that the hyperparameter settings for other datasets are the same as those in our previous experiments on the CIFAR100 dataset, including the learning rate, batch size, and number of training epochs. By achieving consistent excellent results on four different datasets, the ED method demonstrates its reliability and effectiveness in various application scenarios.

\begin{table}
\centering
\caption{Performance of Knowledge Distillation Methods on the Tiny Imagenet Dataset .}\label{tab7}
\begin{tabular}{|c|c|c|}
\hline
Teacher&  ResNet56 & VGG13 \\
Student &   MobileNetV2 & VGG8 \\
\hline
KD & 59.14 & 62.12\\
AT\cite{ref_lncs1}& 57.82& 59.22\\
FitNet\cite{ref_lncs3} & 61.04&62.34 \\
VID\cite{ref_lncs4} &58.33 & 59.05\\
AVER\cite{ref_lncs5} & 57.63 & 62.54  \\
EBKD\cite{ref_lncs6} & 57.88  & 62.76  \\
AEKD\cite{ref_lncs7} & 57.57 &  62.84  \\
\textbf{ED} & \textbf{61.32} & \textbf{61.43}  \\
\hline
\end{tabular}
\end{table}

\begin{table}
\centering
\caption{Performance of Knowledge Distillation Methods on the Caltech256 Dataset and the Food-101 Dataset.}\label{tab8}
\begin{tabular}{|c|c|c|}
\hline
Teacher&   \multicolumn{2}{c|}{ResNet110}\\
Student &    \multicolumn{2}{c|}{ResNet18} \\
Datesets&  Caltech256 & Food-101 \\
\hline
\textbf{ED} & \textbf{53.57} & \textbf{60.69}  \\
\hline
\end{tabular}
\end{table}

\section{Conclusion}
This paper proposes the educational distillation framework, which is inspired by the structured progressive process of human education. By dividing the student model into a main body and a teaching reference block, ED simulates the hierarchical learning stages. It allows the student network to gradually absorb knowledge through sub-datasets while keeping the final architecture unchanged.
ED addresses the problem of catastrophic forgetting by freezing the historical teaching reference blocks and only updating the new modules with the corresponding sub-datasets. Experiments on CIFAR100, Tiny Imagenet, Caltech and Food-101 datasets show that, compared with traditional methods, ED validates its effectiveness in sequential knowledge retention.
The TR block is composed of a student network module, a 1×1 convolutional layer, and an adaptive pooling layer. During training, it serves as a temporary decoder and finally reverts to the original structure. This design avoids architecture distortion and improves training efficiency. In the combination of homogeneous ResNet20/56, the accuracy of ED is 2.2\% higher than that of the multi-teacher knowledge distillation . In the transfer between heterogeneous WRN40 and ShuffleV1, the accuracy reaches 77.63\%, surpassing existing methods.
Although ED performs excellently in fixed stage settings, exploring dynamic stage division (such as adapting to data complexity) and integrating self-supervised learning can further enhance its lifelong learning ability. In addition, extending ED to vision-language models or 3D tasks is expected to open up new application scenarios.
In conclusion, ED bridges the gap between human educational intuition and neural network training, providing a simple and effective solution for knowledge retention and distillation efficiency.

%
% the environments 'definition', 'lemma', 'proposition', 'corollary',
% 'remark', and 'example' are defined in the LLNCS documentclass as well.
%


\end{document}